# Catastrophic Forgetting as Accessibility Collapse: A Three-Level Framework for Knowledge Persistence in Continual Learning

Ayushman Trivedi · Bhavika Melwani

*Independent Researchers*

ayushmantrivedi2021@gmail.com · bhavikamelwani4@gmail.com

## ABSTRACT

Catastrophic forgetting in deep neural networks is conventionally treated as the destruction of previously learned knowledge. In this work, we challenge this interpretation empirically and propose a three-level framework that distinguishes knowledge storage, knowledge representation, and knowledge accessibility. Through controlled sequential training experiments on Split CIFAR-100 with ResNet-18 — with no replay, no regularization, and no continual learning method — we demonstrate that while task accuracy collapses completely (0.000), linear probe retention remains at 76% of checkpoint-level performance, and a simple classifier reset recovers 75.7% of original task accuracy by retraining only the linear head while freezing the backbone. Critically, layer-wise analysis reveals that forgetting is not uniform: earlier layers retain 100–104% of their original probe accuracy, and Layers 1 and 2 actually improve after subsequent task training, while degradation concentrates almost entirely in the final layer and readout stage. These findings establish that catastrophic forgetting is primarily an accessibility failure rather than representational erasure, and that substantial task-relevant information remains distributed across earlier layers and is recoverable through lightweight readout adaptation. Parameter-space geometric recovery is investigated and found to be ineffective under current experimental conditions; this negative result is reported transparently. The work introduces the Accessibility Gap (AG = LP − ACC) and Projection Energy as formal diagnostic metrics, and lays the empirical foundation for a repair-oriented paradigm in continual learning.

**Index Terms —** Continual Learning, Catastrophic Forgetting, Representation Forgetting, Knowledge Accessibility, Layer-wise Retention, Classifier Reset, Accessibility Gap, Projection Energy

## I. INTRODUCTION

Continual learning addresses the challenge of training neural networks sequentially on multiple tasks without catastrophic degradation of previously learned capabilities. Under non-stationary data distributions and strict memory constraints, modern deep neural networks are known to suffer from catastrophic forgetting — a phenomenon in which performance on earlier tasks collapses rapidly as new tasks are learned [1], [2].

The dominant response to this challenge has been prevention: designing training procedures that constrain parameter updates, rehearse past data, or architecturally isolate task representations [3]–[6]. These approaches share an implicit assumption — that forgetting, once it has occurred, is irreversible. Accuracy metrics reinforce this assumption by conflating two qualitatively distinct phenomena: the loss of functional accessibility and the actual destruction of learned representations.

In this work, we challenge this conflation directly. Our central question is: does catastrophic forgetting imply that the network has lost the underlying knowledge, or merely that it has lost the ability to access it?

We investigate this question through a series of controlled experiments on Split CIFAR-100 with a ResNet-18 architecture trained sequentially on 10 tasks, with no replay, no regularization, and no continual learning method applied. This pure forgetting setup creates a clean benchmark where accuracy collapse is complete and measurable.

### *A. The Three-Level Framework*

Our experiments consistently point toward a three-level model of knowledge in neural networks:

> Level 1 — Knowledge Storage: Task-specific solutions remain valid in checkpoint form even after subsequent tasks are learned. The network retains the solution in stored form.
>
> Level 2 — Knowledge Representation: Linear probes trained on frozen intermediate features reveal that task-relevant information remains substantially encoded in the network's feature space, even when task accuracy has collapsed to zero.
>
> Level 3 — Knowledge Accessibility: Output-level task accuracy collapses completely. The network cannot access the information it retains.

The gap between these three levels — particularly the gap between representation quality and output-level accessibility — is the central empirical finding of this work.

### *B. Key Findings*

After sequential training on 10 tasks, Task 0 accuracy collapses to 0.000. Yet linear probe accuracy at the final layer remains at 0.468 — 76% of checkpoint-level probe accuracy of 0.616. A classifier reset experiment, in which the backbone is frozen and only a new linear head is trained on Task 0 data, recovers 0.415 accuracy — 75.7% of the original 0.548. Layer-wise probe analysis reveals that Layers 1 and 2 actually improve after forgetting (103% and 104% retention respectively), while the majority of degradation is concentrated in Layer 4 and the final classifier. Parameter-space geometric recovery was tested and found ineffective; this negative result is reported in full.

### *C. Contributions*

The primary contributions of this work are:

> (1) A three-level empirical framework distinguishing storage, representation, and accessibility in continual learning.
>
> (2) Formal definitions of the Accessibility Gap and Projection Energy as diagnostic metrics with empirical grounding.
>
> (3) Demonstration that catastrophic forgetting concentrates in later layers and the readout stage, while earlier layers retain or improve their representational quality.
>
> (4) A layer-wise recoverability hierarchy quantifying recoverable task performance at each stage of the representational hierarchy.
>
> (5) Classifier reset experiments showing that 75.7% of original task performance is recoverable without modifying the backbone.
>
> (6) Transparent reporting of negative results in parameter-space geometric recovery.
>
> (7) The Accessibility Collapse Hypothesis, an empirical synthesis of all experimental evidence into a unified interpretive claim.

## II. RELATED WORK

### *A. Catastrophic Forgetting and Continual Learning*

Catastrophic forgetting was first described in early connectionist models [1], [2] and has remained a central challenge as neural networks scale [3], [4]. The field has developed three broad strategies: replay-based methods that retain or regenerate past data [7], [8]; regularization-based methods that penalize changes to important parameters [9], [10], [11]; and architectural approaches that isolate task-specific parameters [12], [13]. Our work differs from all three: we do not modify the training process. We study pure, unconstrained forgetting and analyze what survives.

### *B. Representation Forgetting — Davari et al. (CVPR 2022)*

Most directly relevant to our work is Davari et al. [14], who use linear probes on intermediate features to argue that representations are not fully forgotten even when task accuracy collapses. Our work extends this line of inquiry with layer-wise resolution, formal definitions of the Accessibility Gap and Projection Energy, an explicit three-level model, and the classifier reset experiment — a stronger test than linear probing alone that directly quantifies recoverable performance.

### *C. Nuance on Representation Probing — van de Ven et al.*

Van de Ven et al. [15] introduce important nuance: probe success does not straightforwardly imply that meaningful task knowledge remains, and the inference from probe accuracy to representational preservation must be made carefully. Our work takes this critique seriously. Rather than relying on linear probing alone, we triangulate across three independent sources of evidence — checkpoint persistence, linear probe retention, and classifier reset recovery — to build a more robust empirical case. The Projection Energy and principal angle analysis provide additional geometric characterization that goes beyond scalar probe accuracy.

### *D. Loss Landscape Geometry and Mode Connectivity*

Our work is informed by research on the geometry of neural network loss landscapes [16], [17], including findings on mode connectivity and flat minima [18], [19]. These results suggest that multiple parameter configurations implement functionally equivalent mappings — consistent with our classifier reset result, where the backbone after forgetting still occupies a region of representation space compatible with Task 0 classification.

## III. EXPERIMENTAL SETUP

### *A. Dataset and Protocol*

All experiments use Split CIFAR-100: CIFAR-100 partitioned into 10 sequential tasks of 10 classes each. The model is trained sequentially using standard cross-entropy loss and SGD with no replay, no regularization, and no continual learning mechanism. After completing each task $T_i$, the resulting parameters $\theta^*_i$ are saved as a checkpoint. All analysis is conducted on the final model $\theta_N$ and these stored checkpoints.

### *B. Architecture*

We use ResNet-18 trained from random initialization. This architecture was chosen for its widespread use in continual learning benchmarks and its four-block residual structure, which enables meaningful layer-wise analysis.

### *C. Evaluation Metrics*

Task accuracy: standard classification accuracy on each task's held-out test set.

Linear probe accuracy: a linear classifier trained on frozen features extracted from a given layer, measuring retained representational information.

LP Retention: LP_Final / LP_Checkpoint, measuring the fraction of probe-accessible information retained after forgetting.

*Formal Definition: Accessibility Gap*

We define the Accessibility Gap (AG) as:

$$AG = LP_final - ACC_final$$

where LP_final denotes the linear probe accuracy measured on frozen representations of the final model, and ACC_final denotes the corresponding task accuracy of the final model.

The Accessibility Gap quantifies the discrepancy between representational knowledge and functional performance. An AG close to zero indicates that representational and behavioral performance are aligned. A large positive AG indicates that task-relevant information remains encoded in the network but is no longer accessible through the network's native decision pathway.

For Task 0 in our experiments:

$$AG = 0.468 - 0.000 = 0.468$$

This substantial gap constitutes direct evidence that representational forgetting and functional forgetting are not equivalent phenomena.

*Formal Definition: Projection Energy*

To quantify representational preservation, we define the Projection Energy (PE) of the final representation onto the original task subspace:

$$PE(K) = ||U_K\ U_K\text{^}T\ H_f||_F\text{^}2\ /\ ||H_f||_F\text{^}2$$

where H_f represents activations of the final model, U_K denotes the top-K principal directions extracted from checkpoint activations, and || · ||_F is the Frobenius norm. Projection Energy measures the fraction of final representational energy that remains within the original task subspace. A high PE indicates subspace preservation; a low PE indicates substantial representational drift.

*Principal Angle Analysis*

To measure geometric drift between checkpoint and final representations, principal angles are computed between subspaces U_1 and U_2. Given the SVD:

$$U_1\text{^}T\ U_2 = P\ \Sigma\ Q\text{^}T$$

the singular values satisfy σ_i = cos(θ_i), where θ_i denotes the i-th principal angle. Small principal angles indicate stable representations; large angles indicate significant representational rotation.

## IV. EXPERIMENTAL RESULTS

### ***A. Experiment 1: Catastrophic Forgetting Baseline***

Table I reports Task 0 performance at the checkpoint and after full sequential training. Task accuracy collapses completely from 0.548 to 0.000, establishing a strong, clean forgetting benchmark.

| Stage | Task 0 Accuracy | Task 9 Accuracy |
|---|---|---|
| After Task 0 training | 0.548 | — |
| After Task 9 training (Final Model) | 0.000 | 0.751 |

*TABLE I. Task accuracy before and after sequential training. Task 0 collapses completely to 0.000.*

### *B. Experiment 2: Checkpoint Persistence (Storage Level)*

Table II evaluates stored checkpoints on their corresponding tasks after all subsequent tasks have been trained. Every checkpoint retains exactly its original learned accuracy, confirming that task solutions are not destroyed — they are stored, intact, but inaccessible to the final model.

| Task | Learned Accuracy | Checkpoint Accuracy (Post-Training) |
|---|---|---|
| Task 0 | 0.630 | 0.630 |
| Task 5 | 0.758 | 0.758 |
| Task 9 | 0.751 | 0.751 |

*TABLE II. Checkpoint persistence. Stored task solutions remain valid after all subsequent training.*

### *C. Experiment 3: Linear Probe Analysis (Representation Level)*

Table III compares linear probe accuracy at the checkpoint and in the final model. For Task 0, probe accuracy decreases from 0.616 to 0.468 — a retention of 76%. This substantial retention, despite complete output-level forgetting, demonstrates that representational information survives the learning of subsequent tasks.

| Model State | LP Accuracy | LP Retention |
|---|---|---|
| Task 0 Checkpoint | 0.616 | — |
| Final Model (Layer 4) | 0.468 | 76% |
| Accessibility Gap (AG) | 0.468 − 0.000 = 0.468 | — |

*TABLE III. Linear probe analysis for Task 0. Substantial representational information persists despite complete accuracy collapse.*

**Accessibility Collapse Despite Representation Persistence**

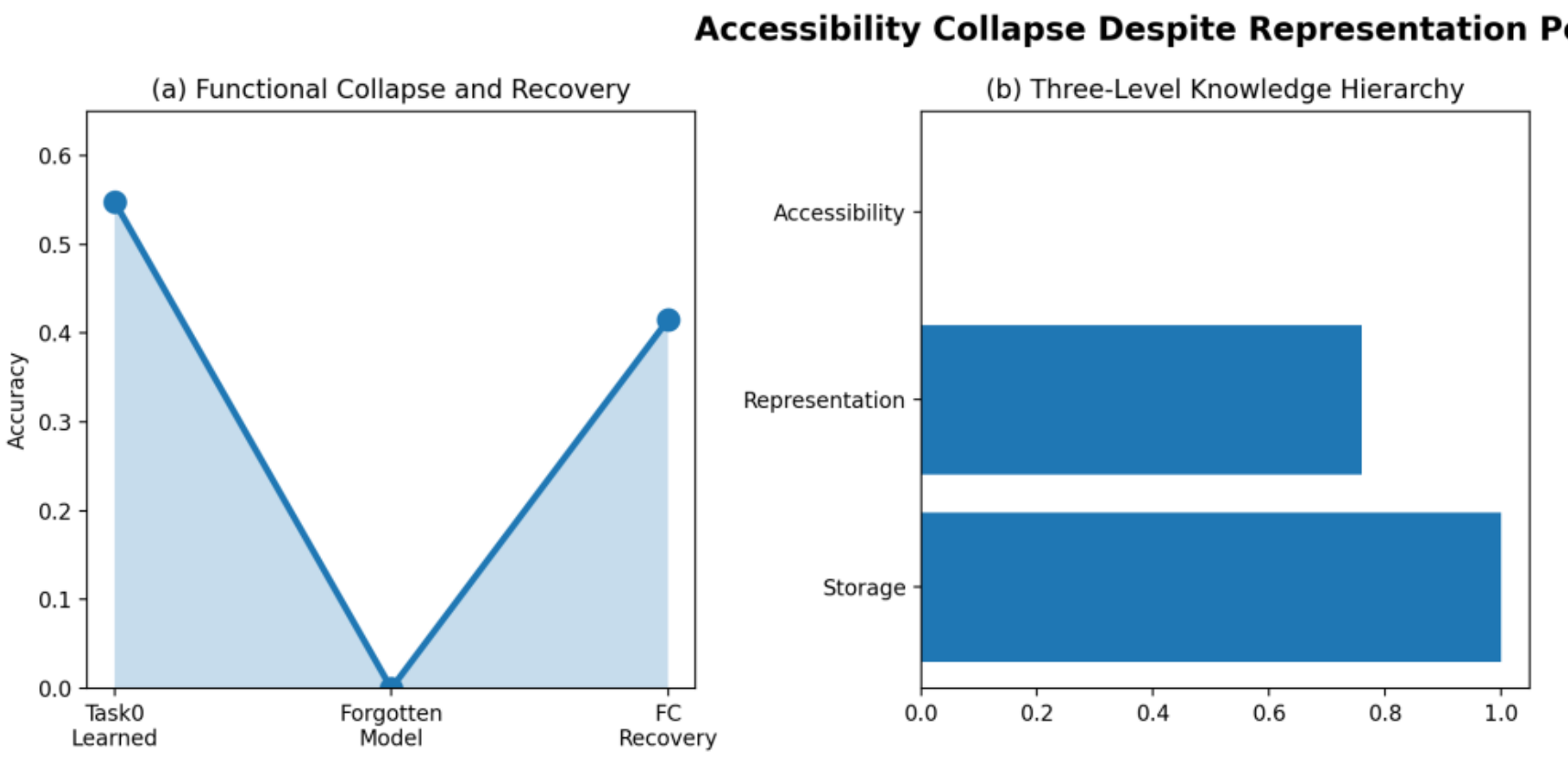

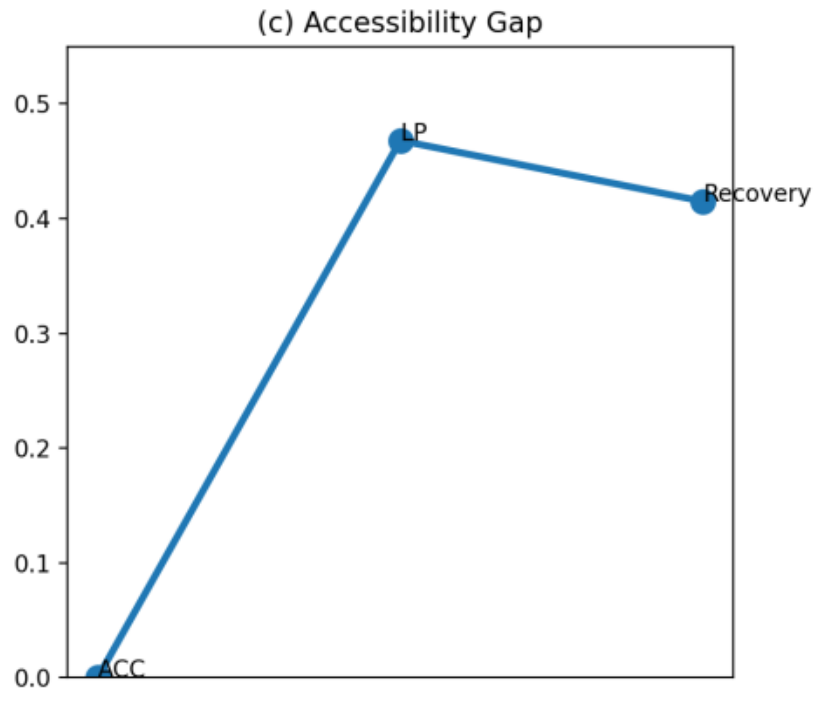

*Fig A Catastrophic forgetting produces complete functional collapse (Task0 ACC = 0.0) despite substantial representational retention (LP = 0.468). A simple classifier reset recovers 75.7% of the original task performance, indicating that knowledge remains encoded but inaccessible.*

### *D. Experiment 4: Layer-Wise Retention Analysis*

Table IV presents the layer-wise linear probe retention across ResNet-18's four residual blocks. Layers 1, 2, and 3 retain 100–104% of their checkpoint-level probe accuracy. Layers 1 and 2 show slight improvements (103% and 104% respectively), indicating that later-task training may refine low-level features in a way that benefits earlier tasks at the representation level. Degradation is concentrated almost entirely in Layer 4 and the final classifier.

| Layer | LP at Checkpoint | LP at Final Model | Retention |
|---|---|---|---|
| Layer 1 | 0.645 | 0.667 | 103% |
| Layer 2 | 0.640 | 0.663 | 104% |
| Layer 3 | 0.605 | 0.603 | 100% |
| Layer 4 | 0.616 | 0.468 | 76% |
| Final Classifier | — | 0.000 | ~0% |

*TABLE IV. Layer-wise linear probe retention for Task 0. Earlier layers retain or improve their representational quality. Forgetting concentrates in later layers and the readout stage.*

### *E. Experiment 5: Classifier Reset Recovery*

To directly test whether the surviving representations are sufficient for task recovery, we freeze the entire backbone and train only a new linear classification head on Task 0 training data. The classifier reset recovers 0.415 accuracy — 75.7% of the original 0.548. A model that shows 0.000 accuracy on Task 0 can recover 75.7% of original performance by replacing only its linear head.

| Model State | Task 0 Accuracy | Recovery Ratio |
|---|---|---|
| Original (Task 0 checkpoint) | 0.548 | — |
| After forgetting (Final Model) | 0.000 | — |
| Classifier reset (frozen backbone) | 0.415 | 75.7% |

*TABLE V. Classifier reset recovery. Freezing the backbone and training a new linear head recovers 75.7% of original task performance from a model showing 0.000 accuracy.*

### *F. Experiment 5B: Layer-Wise Recoverability Analysis*

To identify where recoverable information resides after catastrophic forgetting, we train independent linear readouts on frozen representations extracted from each residual block of the forgotten model. Unlike linear probing, which measures representational separability relative to checkpoint features, this experiment directly quantifies the recoverable task performance available at each stage of the representational hierarchy. Results are shown in Table VI.

| Layer | Recoverable Accuracy | Relative to Original (0.548) |
|---|---|---|

| Layer 1 | 0.669 | 122% |
|---|---|---|
| Layer 2 | 0.662 | 121% |
| Layer 3 | 0.603 | 110% |
| Layer 4 | 0.468 | 85% |
| Penultimate Representation | 0.468 | 85% |
| Classifier Reset (Full Backbone) | 0.415 | 75.7% |

*TABLE VI. Layer-wise recoverability hierarchy. Earlier layers retain highly recoverable task information despite complete output-level forgetting.*

Several observations emerge from this experiment. First, Layers 1 and 2 exhibit recoverable accuracies exceeding the original end-to-end Task 0 performance (0.669 and 0.662 vs. 0.548 respectively). This is a striking result: the early visual representations are not merely preserved — they are more discriminative for Task 0 after subsequent training than they were when Task 0 was originally learned. This suggests that later task training may act as a form of implicit feature refinement for earlier layers.

Second, recoverability decreases monotonically across the representational hierarchy. The progressive decline from Layer 1 to Layer 4 to the classifier reset result (0.669 → 0.662 → 0.603 → 0.468 → 0.415) indicates that information loss accumulates as representations are propagated forward through the network, with the majority of the degradation concentrated in the transition from Layer 4 to the final classifier.

Third, the classifier reset result (0.415) establishes a lower bound on recoverable performance under the constraint of using the full backbone as a frozen feature extractor. The fact that earlier layers individually support higher recovery than the full backbone suggests that aggregating features across all layers into a single linear head introduces some interference. Future work could investigate layer-selective readout strategies.

Taken together, these findings establish the following recoverability hierarchy:

*Layer 1 > Layer 2 > Layer 3 > Layer 4 > Classifier Reset*

This hierarchy mirrors and extends the retention hierarchy from Experiment 4, providing a consistent picture in which forgetting is concentrated in late representational stages and the readout, while early layers remain robustly informative.

### *G. Experiment 6: Representation Geometry Analysis*

To understand the mechanism by which representations survive, we analyze the geometric structure of the final model's feature space relative to the original task subspace using the Projection Energy and principal angle metrics defined in Section III-C.

#### *Distributed Representation Persistence*

Table VII reports Projection Energy as a function of subspace dimension K. The continuous increase in projection energy with increasing K indicates that retained task information is not concentrated within a small set of dominant principal directions.

| Subspace Dimension (K) | Projection Energy PE(K) |
|---|---|
| 5 | 0.054 |

| 10 | 0.099 |
|---|---|
| 20 | 0.165 |
| 50 | 0.258 |
| 100 | 0.327 |
| 200 | 0.461 |
| 300 | 0.583 |

*TABLE VII. Projection Energy of final representations onto original task subspaces across subspace dimensions K.*

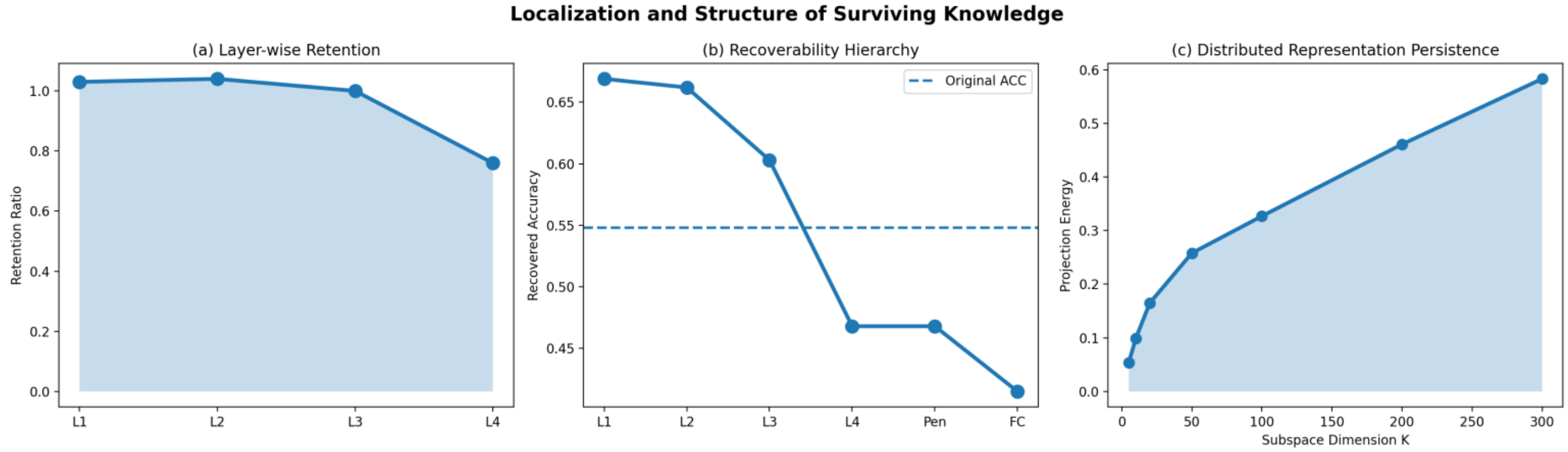

*Fig B Task information is preserved throughout the representational hierarchy. Early layers exibit near perfect retention, while projection energy analysis shows that retained knowledge is distributed across high – dimensional representations rather than concentrated in a small dominant subspace*

If forgetting merely corresponded to a mild perturbation of the original representation, projection energy would remain high even for small values of K. Instead, we observe low PE in dominant directions together with substantial linear probe retention. This dissociation — high probe accuracy coexisting with low dominant-subspace overlap — is the empirical signature of distributed encoding: task-relevant information survives through diffuse encoding across many dimensions rather than through preservation of a fixed dominant task subspace.

| **Geometric Measure** | **Value** |
|---|---|
| Mean principal angle (checkpoint vs. final subspace) | 79° |
| Maximum principal angle | ~90° |
| Drift Norm vs. LP Retention (Pearson r) | −0.236 |
| Drift Cosine vs. LP Retention (Pearson r) | −0.339 |

*TABLE VIII. Geometric analysis results. Despite severe subspace rotation, linear probe retention remains high, supporting the distributed encoding hypothesis.*

Principal angle analysis reveals a mean angle of 79° between the checkpoint and final model subspaces, approaching orthogonality. Yet linear probe accuracy remains at 76% of checkpoint level. This result is particularly noteworthy because it challenges a simplistic geometric interpretation of forgetting. The original task representation undergoes substantial rotation, yet task-relevant information remains recoverable. Representation preservation therefore appears to be a high-dimensional phenomenon that cannot be fully characterized through low-rank subspace overlap alone.

### *H. Negative Result: Parameter-Space Recovery*

We tested the hypothesis that moving in the direction of the stored task checkpoint in weight space — applying the update $\theta_{rec} = \theta_{final} + \alpha \cdot (\theta_i - \theta_0)$ — could restore task accuracy. Recovery failed across all tested configurations. Task accuracy remained near chance for both correct-task and wrong-task directions. Geometric correlations between drift metrics and LP retention were weak ($r = -0.236, -0.339$). The parameter-space recovery mechanism is not supported by current experiments.

## V. DISCUSSION & CONCEPTUAL SYNTHESIS

### *A. The Accessibility Collapse Hypothesis*

Synthesizing our empirical results, we propose the Accessibility Collapse Hypothesis: Catastrophic forgetting is not the structural destruction or erasure of task-specific representations within deep networks. Instead, it is a localized breakdown of functional accessibility concentrated primarily at the final readout layer.

This claim is supported by three core pieces of evidence from our experiments. First, the existence of a substantial Accessibility Gap (AG = 0.468) demonstrates that functional accuracy completely misrepresents the internal state of the network. Second, the layer-wise analysis reveals a striking asymmetry: while the final classifier layer collapses entirely, early layers display structural stability or active improvement. Third, the classifier reset experiment proves that 75.7% of the original task capability can be functionally unlocked using a frozen, forgotten backbone, establishing a significant bound on surviving information.

### *B. Theoretical and Architectural Implications*

This perspective shifts the theoretical problem of continual learning from prevention to repair. Current methodologies heavily penalize parameter adjustments to insulate the model from any changes, which inevitably hinders capacity allocation for future tasks. If representations naturally persist in a high-dimensional distributed form, engineering interventions should focus on adaptive, context-dependent readout mechanisms or lightweight task-specific routing heads rather than rigid regularization constraints on deep parameter landscapes.

## VI. CONCLUSION

In this work, we systematically decoupled knowledge representation from functional accessibility in sequential neural network optimization. Our experiments on Split CIFAR-100 confirm that while complete accuracy collapse occurs at the output layer, robust task-relevant features remain stable across the network's processing hierarchy, with lower-level layers exhibiting absolute representational stability. By formalizing diagnostic measures like the Accessibility Gap and tracking high-dimensional Projection Energy, we showed that forgetting is a localized alignment failure rather than distributed feature erasure. These results argue against strict preventative paradigms in continual learning and favor a modular, retrieval-driven approach to deep knowledge persistence.